\documentclass[sigconf]{acmart}
\AtBeginDocument{%
  }

\copyrightyear{2026}
\acmYear{2026}
\setcopyright{cc}
\setcctype{by}
\acmConference[KDD '26]{Proceedings of the 32nd ACM SIGKDD Conference on Knowledge Discovery and Data Mining V.2}{August 09--13, 2026}{Jeju Island, Republic of Korea}
\acmBooktitle{Proceedings of the 32nd ACM SIGKDD Conference on Knowledge Discovery and Data Mining V.2 (KDD '26), August 09--13, 2026, Jeju Island, Republic of Korea}
\acmDOI{10.1145/3770855.3818334}
\acmISBN{979-8-4007-2259-2/2026/08}

\begin{document}

\title[CanniUplift for Mitigating Seller and Incentive Cannibalization in E-commerce Uplift Modeling]{CanniUplift: A Holistic Framework for Mitigating Seller and Incentive Cannibalization in E-commerce Uplift Modeling}

\author{Zuwang He}
\authornote{These authors contributed equally to this work.}
\orcid{0009-0009-5223-972X}
\email{zuwanghe.hzw@gmail.com}
\affiliation{%
  \institution{Taobao \& Tmall Group of Alibaba}
  \city{Beijing}
  \country{China}}

\author{Shihao Shu}
\authornotemark[1]
\orcid{0009-0003-0540-2009}
\email{shushihao.ssh@alibaba-inc.com}
\affiliation{%
  \institution{Taobao \& Tmall Group of Alibaba}
  \city{Beijing}
  \country{China}}

\author{Yuli Qu}
\authornotemark[1]
\orcid{0009-0001-1118-9203}
\email{quyuli.qyl@alibaba-inc.com}
\affiliation{%
  \institution{Taobao \& Tmall Group of Alibaba}
  \city{Hangzhou}
  \state{Zhejiang}
  \country{China}}

\author{Hanyu Gao}
\orcid{0009-0008-4319-119X}
\email{gaohanyu.ghy@alibaba-inc.com}
\affiliation{%
  \institution{Taobao \& Tmall Group of Alibaba}
  \city{Beijing}
  \country{China}}

\author{Ziliang Zhang}
\orcid{0009-0004-4807-3652}
\email{zhangziliang.zzl@alibaba-inc.com}
\affiliation{%
  \institution{Taobao \& Tmall Group of Alibaba}
  \city{Beijing}
  \country{China}}

\author{Diwei Chen}
\orcid{0000-0001-6080-9969}
\email{chendiwei.cdw@alibaba-inc.com}
\affiliation{%
  \institution{Taobao \& Tmall Group of Alibaba}
  \city{Hangzhou}
  \state{Zhejiang}
  \country{China}}

\author{Xiangda Yan}
\orcid{0009-0009-5952-5139}
\email{xiangda.yxd@alibaba-inc.com}
\affiliation{%
  \institution{Taobao \& Tmall Group of Alibaba}
  \city{Hangzhou}
  \state{Zhejiang}
  \country{China}}

\author{Buyu Gao}
\orcid{0009-0004-9686-458X}
\email{gaobuyu.gby@alibaba-inc.com}
\affiliation{%
  \institution{Taobao \& Tmall Group of Alibaba}
  \city{Hangzhou}
  \state{Zhejiang}
  \country{China}}

\author{Tanchao Zhu}
\orcid{0009-0003-6474-0868}
\email{tanchao.zhutc@alibaba-inc.com}
\affiliation{%
  \institution{Taobao \& Tmall Group of Alibaba}
  \city{Hangzhou}
  \state{Zhejiang}
  \country{China}}

\author{Yumeng Li}
\authornote{Corresponding author.}
\orcid{0009-0000-0801-6398}
\email{lym174806@alibaba-inc.com}
\affiliation{%
  \institution{Taobao \& Tmall Group of Alibaba}
  \city{Beijing}
  \country{China}}

\author{Junxiong Zhu}
\orcid{0009-0003-4350-5324}
\email{xike.zjx@alibaba-inc.com}
\affiliation{%
  \institution{Taobao \& Tmall Group of Alibaba}
  \city{Hangzhou}
  \state{Zhejiang}
  \country{China}}

\renewcommand{\shortauthors}{Zuwang He et al.}

\begin{abstract}
  Personalized incentive allocation is vital for e-commerce, where uplift modeling is the standard for estimating Individual Treatment Effects (ITE). However, traditional models often fail in complex multi-seller environments with violations of the Stable Unit Treatment Value Assumption (SUTVA). We identify two critical challenges: Seller-level Cannibalization, where incentives shift expenditure between shops without growing the platform, and Incentive-level Cannibalization, where organic conversions or alternative rewards introduce significant noise into incrementality estimation. In this paper, we propose CanniUplift, a unified framework to mitigate these dual-source cannibalization effects. Specifically, we design Platform-level Global Alignment (PGA) to capture cross-shop substitution through global GMV consistency constraints. To tackle incentive-driven noise, we introduce Redemption-based Decomposition Denoising (RDD), which uses redemption behavior to decompose treated outcomes and reduce attribution noise within an entire-space framework. Furthermore, a Treat-Attention mechanism is designed to model intricate interactions between users' historical behaviors and current treatment options. Extensive experiments on both synthetic and large-scale industrial datasets demonstrate that CanniUplift significantly outperforms state-of-the-art baselines. Ablation studies confirm that the integration of PGA and RDD consistently improves wAUUC and wQINI. Successfully deployed online, our framework achieved a 4.08\% relative increase in platform-wide incremental GMV ($\Delta$GMV) over the production baseline and improved ROI in online A/B tests, proving effective in driving global platform growth.
\end{abstract}





\maketitle

\section{Introduction}

\begin{figure}[htbp]
\centering
\includegraphics[width=0.48\textwidth]{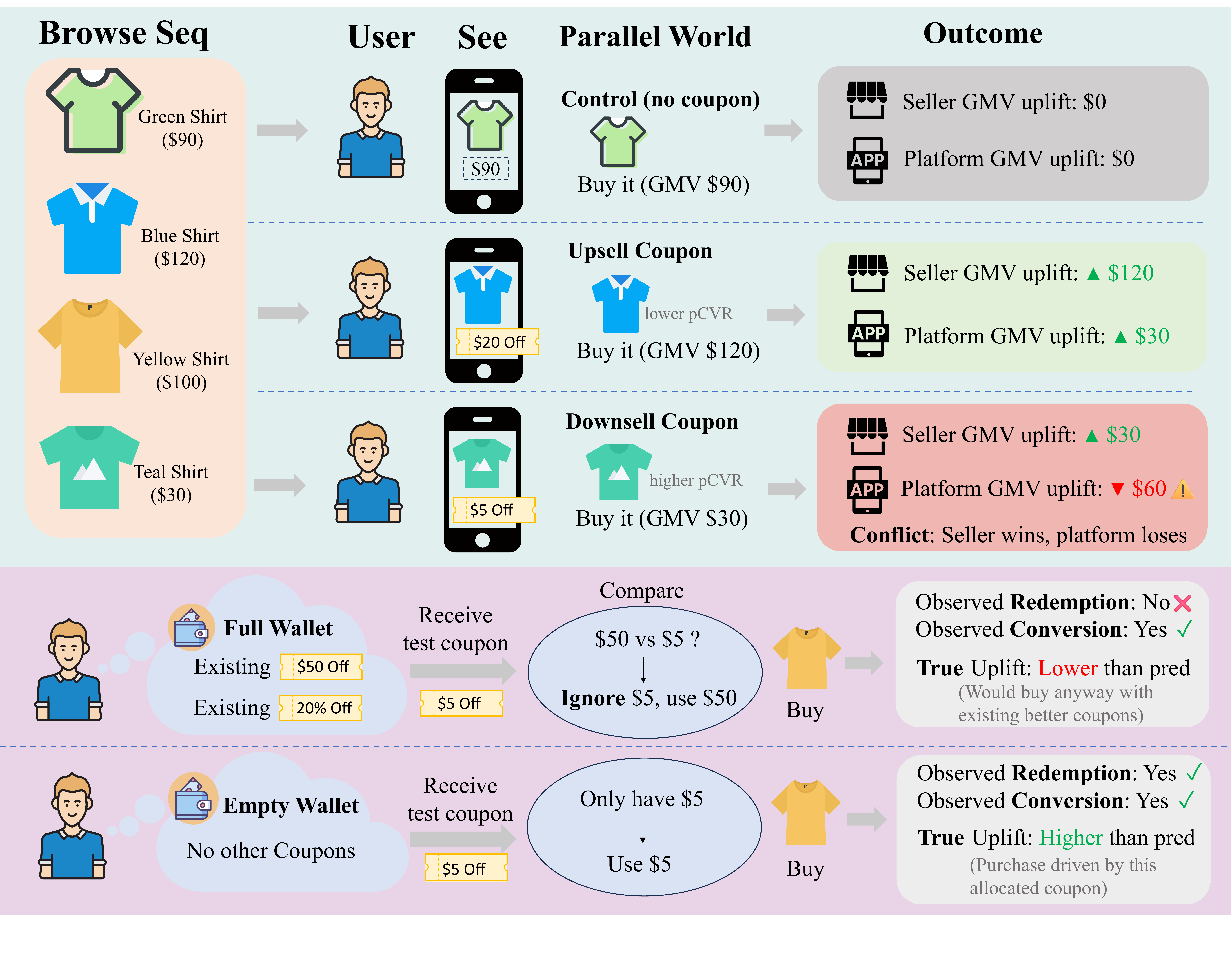}
\caption{Two critical violations of SUTVA in multi-seller e-commerce ecosystems. \textbf{Top:} Seller-level cannibalization. \textbf{Bottom:} Incentive-level cannibalization.}
\label{fig:intro_fig_0207}
\end{figure}

With the rapid proliferation of digital commerce, personalized marketing has become a pivotal strategy for e-commerce platforms to drive user engagement and revenue growth. At the heart of these strategies lies the allocation of incentives, such as digital coupons and subsidies. Unlike traditional conversion rate (CVR) models that focus on predicting user behavior, Uplift Modeling has emerged as the standard for estimating the Individual Treatment Effect (ITE)—the true incremental value driven by a specific intervention. By identifying ``persuadable'' users, platforms can optimize resource allocation and maximize overall returns.
However, applying uplift modeling in a complex, multi-seller e-commerce ecosystem faces significant theoretical and practical hurdles. Most existing uplift frameworks, such as DragonNet and EUEN, implicitly rely on the Stable Unit Treatment Value Assumption (SUTVA). This assumption posits that the treatment assigned to one unit does not affect the outcomes of other units. In a large-scale marketplace, this assumption is frequently violated due to what we term Multi-source Cannibalization Effects, as illustrated in Fig.~\ref{fig:intro_fig_0207}, which manifest in two critical dimensions:

First, Seller-level Cannibalization. Traditional uplift models optimize for individual seller increments, which inadvertently triggers ``downsell effects''. As shown in Fig.~\ref{fig:intro_fig_0207} (top), a user might originally intend to purchase a \$90 item, but a \$5 coupon allocated to a different seller nudges them toward a \$30 alternative. While the individual seller records a gain, the platform GMV drops. Such models fail to capture this ``zero-sum'' substitution, where seller uplift is achieved at the expense of global platform growth.

Second, Incentive-level Cannibalization. In complex e-commerce environments, users are often exposed to multiple concurrent rewards. As shown in Fig.~\ref{fig:intro_fig_0207} (bottom), users with existing higher-value coupons may ignore the assigned test coupon but still complete a transaction. This ``conversion without redemption'' behavior leads to biased uplift estimation and wastes allocation opportunity costs, as the model misattributes organic motivation or alternative incentives to the assigned treatment.

We propose CanniUplift, a holistic framework to mitigate dual-source cannibalization and boost platform-wide incremental GMV. Our contributions are four-fold: 

\begin{itemize}

\item We relax the SUTVA constraint via a multi-head architecture with a global GMV objective. By aggregating shop-level predictions under holistic constraints, the model implicitly learns cross-seller substitution effects, ensuring optimization aligns with platform-wide growth.

\item We propose a redemption-based decomposition module. By separating conversions with and without redemption, the model reduces attribution noise from organic demand and concurrent incentives, leading to more robust uplift estimation.

\item We design a treat-attention mechanism that models the intricate interactions between a user's historical item-level engagement and their responsiveness to different incentive types, providing a more granular representation for personalized allocation.

\item Extensive evaluations on industrial and synthetic datasets show CanniUplift consistently outperforms state-of-the-art baselines. Successfully deployed on a real-world e-commerce system, the framework delivers marked gains in platform-wide incremental GMV and ROI.

\end{itemize}

\section{Preliminary}

In this section, we establish the mathematical notation and foundational framework for uplift modeling within the context of platform-wide e-commerce marketing. Unlike traditional supervised learning, uplift modeling aims to estimate the causal effect of a specific intervention.

\subsection{Problem Setup and Notations}
Let $\mathcal{U}$ be the set of users and $\mathcal{S}$ be the set of sellers (shops) on an e-commerce platform. For a marketing campaign, we observe a dataset:
\begin{equation}
\mathcal{D} = \{(x_i, t_{i,s}, r_{i,s}, y_{i,s})\}_{i=1}^N,
\end{equation}

where $x_i \in \mathbb{R}^d$ represents the feature vector of user $i$, encompassing historical behaviors and context, $t_{i,s} \in \{0, 1\}$ is the treatment indicator, where $t_{i,s}=1$ if user $i$ is assigned an incentive (e.g., a shop-specific coupon) for seller $s$, and $t_{i,s}=0$ otherwise, $r_{i,s} \in \{0, 1\}$ is the redemption indicator, representing whether user $i$ actually utilized the assigned incentive for the transaction, and $y_{i,s} \in \mathbb{R}_{\geq 0}$ denotes the observed outcome, such as the Gross Merchandise Volume (GMV) contributed by user $i$ to seller $s$.

\subsection{The Potential Outcome Framework}
Following the Neyman-Rubin Potential Outcome Framework \cite{rubin1974estimating}, each user $i$ has two potential outcomes for a given seller $s$: $y_{i,s}(1)$ and $y_{i,s}(0)$, representing the outcome if the user were treated or not, respectively. The \textit{Individual Treatment Effect} (ITE), also known as Conditional Average Treatment Effect (CATE), is defined as:
\begin{equation}
\tau_{i,s}(x_i) = \mathbb{E}[y_{i,s}(1) - y_{i,s}(0) | X=x_i].
\end{equation}
The fundamental challenge of causal inference is that only one potential outcome is observed for any individual, i.e., $y_{i,s} = t_{i,s} y_{i,s}(1) + (1-t_{i,s}) y_{i,s}(0)$.

\subsection{SUTVA and the Cannibalization Challenge}

Standard uplift models typically rely on the \textbf{Stable Unit Treatment Value Assumption (SUTVA)} \cite{holland1986statistics}. In practice, SUTVA is commonly interpreted as:
\begin{enumerate}
    \item \textbf{No Interference}: the potential outcome of a unit is unaffected by the treatment assignments of other units.
    \item \textbf{No Hidden Versions of Treatment}: the treatment is well-defined, i.e., there are no multiple versions of the ``same'' treatment that would lead to different potential outcomes.
\end{enumerate}

In our scenario, SUTVA is frequently violated due to \textbf{Seller Cannibalization}. Specifically, an incentive for seller $A$ may divert a user's organic expenditure from seller $B$. Thus, the potential outcome $y_{i,s}$ is not independent of treatments assigned to other sellers $\{t_{i,s'} \}_{s' \neq s}$. To address this, we define the \textbf{Global Platform GMV} for user $i$ as:
\begin{equation}
Y_i = \sum_{s \in \mathcal{S}} y_{i,s}.
\end{equation}

Rather than merely optimizing shop-level metrics, our objective is to maximize the global incrementality $\Delta Y_i = \mathbb{E}[Y_i | \text{Treat}] - \mathbb{E}[Y_i | \text{Control}]$.

\subsection{Incentive-level Noise}
In real-world marketing, a user may be assigned a coupon ($t_{i,s}=1$) but purchase without redeeming it ($r_{i,s}=0$), a \emph{noncompliance} phenomenon where the observed outcome is a mixture of latent compliance types. Formally, we regard $r$ as a post-treatment variable, decomposing the observed outcome distribution as:
\begin{equation}
\begin{aligned}
p(y \mid t=1)
&= p(r=1 \mid t=1)\, p(y \mid t=1, r=1) \\
&\quad + p(r=0 \mid t=1)\, p(y \mid t=1, r=0).
\end{aligned}
\end{equation}
The second component ($t=1, r=0$) corresponds to purchases that happen without redeeming the assigned coupon, which may be driven by organic demand or other concurrent incentives. This component contaminates uplift estimation if one naively treats all $t=1$ conversions as incremental. Therefore, it is crucial to distinguish the ``conversion with redemption'' and ``conversion without redemption'' mechanisms to reduce attribution bias in estimating the incremental value of the assigned incentive.

\subsection{Formal Definition of Cannibalization Effects}
To mathematically characterize the impact of interventions in a non-SUTVA environment, we define the \textit{Platform Net Increment} $\Delta Y_i(s_j)$ as the total change in user $i$'s platform-wide expenditure $Y_i$ after assigning treatment $t_{i,s_j}=1$ (e.g., a specific shop-coupon) to seller $s_j$:

\begin{equation}
\begin{aligned}
\Delta Y_i(s_j) &= \mathbb{E}[Y_i | t_{i,s_j}=1] - \mathbb{E}[Y_i | t_{i,s_j}=0] \\
&= \sum_{s_k \in \mathcal{S}} \left( \mathbb{E}[y_{i,s_k} | t_{i,s_j}=1] - \mathbb{E}[y_{i,s_k} | t_{i,s_j}=0] \right) \\
&= \underbrace{\left( \mathbb{E}[y_{i,s_j} | t_{i,s_j}=1] - \mathbb{E}[y_{i,s_j} | t_{i,s_j}=0] \right)}_{\text{Apparent Increment (Local Lift)}} \\
&+ \underbrace{\sum_{s_k \neq s_j} \left( \mathbb{E}[y_{i,s_k} | t_{i,s_j}=1] - \mathbb{E}[y_{i,s_k} | t_{i,s_j}=0] \right)}_{\text{Cross Effect: Cannibalization (-) or Spillover (+)}}
\end{aligned}
\end{equation}

where $y_{i,s_k}$ denotes the GMV contributed by user $i$ to seller $s_k$, as defined in Eq. (3). Based on this formulation, we generalize the cannibalization effect into two dimensions:
\begin{itemize}
    \item \textbf{Seller-level Cannibalization:} This occurs when the Cross Effect term in the above equation is negative. It indicates that the observed growth in shop $s_j$ is partially or fully offset by a decrease in other shops $s_k$ ($k \neq j$), as the user merely reallocates their limited budget instead of generating new demand.
    \item \textbf{Incentive-level Cannibalization:} This reflects the interaction between different reward types or organic motivation. When a user who would have converted organically (or via a platform-wide reward) utilizes the assigned shop coupon $t_{i,s_j}=1$, the Apparent Increment term becomes a ``pseudo-incrementality'', effectively cannibalizing the platform's potential profit or organic GMV that would have occurred without the specific treatment.
\end{itemize}

\section{Methodology}

\begin{figure*}[!h]
\centering
\includegraphics[width=\textwidth]{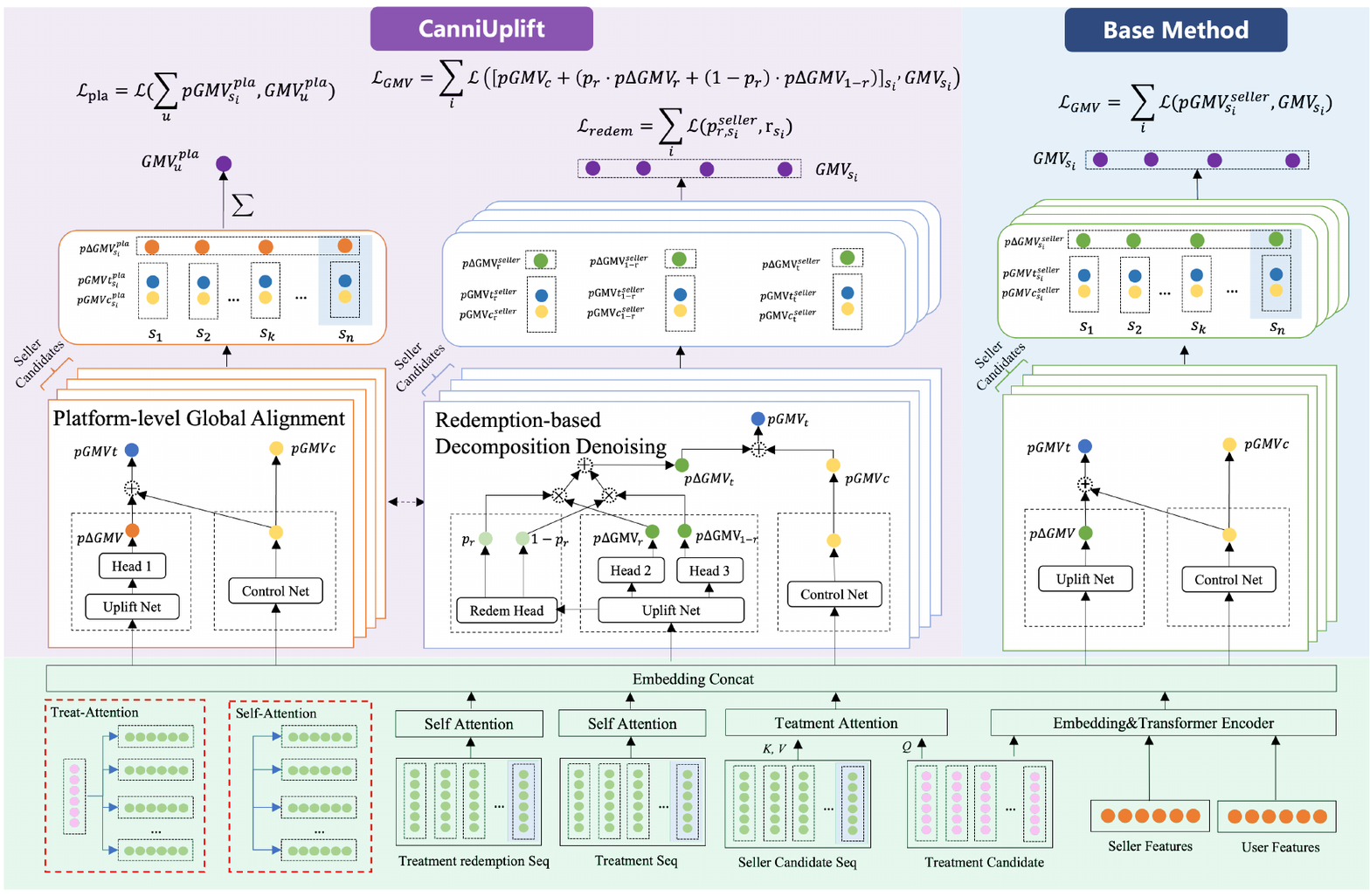}
\caption{The architecture of the CanniUplift framework. The model consists of three core components: (Left) The Platform-level Global Alignment (PGA) module, which applies a global GMV consistency constraint to capture cross-shop substitution effects. (Middle) The Redemption-based Decomposition Denoising (RDD) module, which uses redemption behavior to decompose treated outcomes into redemption and non-redemption paths and reduce incentive-level attribution noise. (Right) The Seller-view Local Uplift module, which provides baseline ITE estimation. All modules are supported by a shared Multi-source Sequence Encoder and a Treat-Attention mechanism that captures fine-grained user-candidate interactions.}
\label{fig:example}
\end{figure*}


Fig.~\ref{fig:example} illustrates the overall framework. It is important to emphasize that the green module on the right corresponds to the \textbf{baseline} structure---a seller-view local uplift learning paradigm, whose core is to model the treated/control outcomes for each candidate seller independently and estimate uplift. In contrast, the left blue PGA module addresses Seller Cannibalization through platform-level global alignment, while the middle purple RDD module addresses Incentive Cannibalization through redemption-based decomposition. All three parts share the same bottom representation and candidate-interaction encoder designed in this work, while we redesign the task heads and training objectives to upgrade uplift learning from \emph{local incremental estimation} to a unified framework of \emph{denoising + global alignment}, which better matches real-world e-commerce environments with multi-seller competition and concurrent incentives.

\subsection{Representation Learning and Candidate Interaction}\label{sec:3-1}
In e-commerce incentive delivery, whether a user will be triggered to generate incremental value by a coupon or a shop largely depends on recent behaviors and historical coupon usage patterns, rather than static profiles alone. More importantly, the heterogeneity of uplift mainly arises from the \emph{user--candidate} interaction: the same user may respond very differently to different shops, coupon amounts, thresholds, or validity periods. Therefore, instead of compressing all signals into a single user vector, CanniUplift adopts a two-stage design---\textbf{multi-source sequence encoding + candidate treat-attention}---to explicitly generate a candidate-specific interaction representation for each candidate.

\paragraph{Multi-source sequence encoding (Shared Encoder).}
We construct three types of inputs, apply embedding, and feed them into a Transformer encoder:

\textbf{Candidate sets} (used by the subsequent treat-attention):

    \begin{enumerate}
        \item treatment candidates: coupon/red-packet attributes including threshold, discount type (discount vs. full-reduction), face value, etc.;
        \item seller candidates: the set of candidate shops that can be delivered/recommended in the current decision.
    \end{enumerate}
    
\textbf{Statistical static features} (encoded by embedding layers): e.g., shop-side category information and user-side profile features.

\textbf{Sequential features} (encoded by self-attention):
    
    \begin{enumerate}
    \item user behavior sequence: time-ordered item/shop behaviors (impression/click/purchase, etc.), capturing short-term purchase intent and preference drift;
    \item treatment sequence: historical incentive exposure /claim /redemption, capturing sensitivity to incentive mechanisms and redemption propensity.
    \end{enumerate}

Multi-source inputs are encoded by a shared embedding layer and a Transformer encoder, producing the sequence representation $H_u$ and the pooled representation $h_u$ for downstream use:
\begin{equation}
H_u = \text{Transformer}\big([E_{\text{beh}}, E_{\text{ben}}, E_{\text{oth}}]\big), \quad
h_u = \text{Pool}(H_u).
\end{equation}
Here, $H_u$ retains fine-grained temporal signals (e.g., ``whether the last purchase redeemed a coupon'' and ``recently preferred shops or categories''), which serve as the Key/Value for candidate interaction.

\paragraph{Candidate interaction encoding: Treat-Attention (Candidate Encoder).}
For each user $u$, the system typically provides a set of candidates (candidate shops, candidate coupons, or their combinations). We denote each candidate by $c$, whose feature vector $E_c$ is formed by concatenating the candidate shop embedding and coupon attributes (face value, threshold, validity period, applicable categories, etc.). To capture heterogeneous responses across candidates, we adopt Treat-Attention: using the candidate as Query and the multi-source sequence representation as Key/Value to obtain a candidate-level interaction representation:
\begin{equation}
z_{u,c} = \text{Attn}\big(q(E_c), K(H_u), V(H_u)\big).
\end{equation}

This mechanism explicitly captures which historical fragments are most relevant to the current candidate. For price-sensitive users, attention may focus more on historical coupon claiming and redemption behaviors; for brand-loyal users, it may concentrate on interactions with the current seller or similar sellers; for highly substitutable products, it may highlight same-category cross-seller browsing, which provides informative signals for learning seller cannibalization. Finally, each candidate obtains an interaction vector $z_{u,c}$, which is fed into the seller-level, platform-level, and redemption-related heads to predict seller GMV, enforce platform-level GMV consistency, and construct the RDD-based denoised GMV prediction.

\subsection{Baseline Structure: Seller-view Local Uplift Learning}
The green module on the right represents a common industrial uplift/ITE learning paradigm: it predicts GMV under treatment/control (or equivalently uplift) for each candidate seller and optimizes with shop-level supervision: \textbf{Control Net} outputs $Y_0$ (expected GMV without coupon) and \textbf{Uplift Net} outputs $Y_1$ (expected GMV with coupon), and computes $\text{uplift}=Y_1-Y_0$. For each candidate seller, its observed GMV is used to supervise the corresponding output.

Such seller-view methods implicitly rely on SUTVA (no interference). In platform marketing, this assumption is often violated due to cross-seller substitution and incentive noncompliance/credit contamination, which motivates our two corresponding modules introduced below.

\subsection{Platform-level Global Alignment}
To directly optimize platform-wide net incrementality rather than single-shop growth, we introduce a \textbf{Platform head} (Head1 in Fig.~\ref{fig:example}) on top of the shared representation and candidate encoder. It aggregates predictions over candidate sellers and aligns them with the user's total platform GMV, encouraging the model to capture cross-seller cannibalization effects from the training objective.

Given the candidate seller set $\mathcal{S}_u$ for user $u$, the platform head outputs an aggregatable prediction $pGMV^{\text{pla}}_{s_i}$ for each $s_i\in\mathcal{S}_u$, and computes:
\begin{equation}
pGMV^{\text{pla}}_u = \sum_{s_i \in \mathcal{S}_u} pGMV^{\text{pla}}_{s_i}.
\end{equation}
From logs we also obtain the user's total platform GMV within the observation window (summed across shops). During training, we minimize their discrepancy:
\begin{equation}
\mathcal{L}_{\text{pla}} = \sum_u \mathcal{L}\left(pGMV^{\text{pla}}_u, GMV^{\text{pla}}_u\right).
\end{equation}
Under this constraint, if the model predicts high uplift for multiple shops simultaneously from the seller view, their aggregation will systematically inflate $pGMV^{\text{pla}}_u$ and be penalized by the platform loss. Conversely, when a shop's ``lift'' mainly comes from other shops' ``loss'', the total platform pie does not grow, and the platform constraint suppresses such \emph{spurious uplift}. Therefore, the platform head provides an \textbf{additive consistency} global training signal, enabling the model to implicitly learn substitution relations without explicitly modeling all pairwise seller interactions.

While training aggregates all sellers per user to enforce global consistency, inference only requires single-seller predictions to estimate platform-level uplift. The detailed derivation is provided in Appendix~\ref{sec:appendix_inference}.

\subsection{Redemption-based Decomposition Denoising}
In practice, treatment assignment (issuing a coupon) does not necessarily imply treatment receipt (redeeming it). A user may purchase under assignment but not redeem the assigned coupon ($r=0$), either due to purely organic motivation or because other concurrent incentives ``hijack'' the conversion credit. This noncompliance phenomenon makes $t{=}1$ outcomes a mixture of heterogeneous mechanisms, which can severely contaminate uplift attribution if one naively treats all treated conversions as incremental.

To explicitly model this incentive cannibalization, we introduce Redemption-based Decomposition Denoising (RDD), which factorizes the treated outcome into two disjoint paths: purchase with redemption and purchase without redemption. For each user--candidate seller pair $(u,s_i)$, the \textbf{Redem Net} predicts the redemption probability:
\begin{equation}
p_r^{s_i} \triangleq p(r=1 \mid z_{u,s_i}),
\end{equation}
where $z_{u,s_i}$ is the candidate-specific interaction representation produced by Treat-Attention. The treatment-side head outputs two path-specific GMV increments, denoted as $p\Delta GMV_r^{s_i}$ and $p\Delta GMV_{1-r}^{s_i}$, corresponding to the redemption path ($r{=}1$) and the non-redemption path ($r{=}0$), respectively. The control-side head outputs the baseline GMV under no treatment, denoted as $pGMV_c^{s_i}$.

Following the law of total expectation, the final seller-level GMV prediction used for supervision is constructed as a redemption-weighted mixture:
\begin{equation}
\widehat{GMV}_{u,s_i}
= pGMV_c^{s_i}
+ p_r^{s_i} \cdot p\Delta GMV_r^{s_i}
+ (1-p_r^{s_i}) \cdot p\Delta GMV_{1-r}^{s_i}.
\end{equation}

This design encourages the model to explain treated outcomes through two behaviorally distinct paths, so that conversions without redemption are not forced to share the same uplift pattern as conversions with redemption. Unlike a pure auxiliary redemption prediction task, RDD embeds redemption directly into the main treated-outcome path, which follows the spirit of \emph{Entire-Space}\cite{ma2018entire} modeling by explicitly accounting for the post-treatment redemption dimension. Here, $\widehat{GMV}_{u,s_i}$ is supervised by the observed seller-level GMV through the seller-level Tweedie loss, while $p_r^{s_i}$ is supervised by the observed redemption label through the redemption classification loss. The formal definitions of these loss terms, together with the platform-level aggregation loss, are given in Sec.~\ref{sec:loss}.

\subsection{Loss Function}
\label{sec:loss}

GMV prediction in e-commerce typically exhibits \textbf{zero inflation} (many users do not purchase, yielding $y=0$) and a \textbf{long-tail distribution} (a small fraction of large orders dominate). To better fit both properties, improve training stability, and enhance tail modeling, we adopt the \textbf{Tweedie loss} for all GMV-related outputs (seller local GMV, platform aggregated GMV, and GMV heads in the denoising paths):
\begin{equation}
\mathcal{L}_{\text{GMV}}^{\text{Tw}} = \sum \mathcal{L}_{\text{Tw}}(y, \hat{y}), \quad (\text{Tweedie family, } 1 < p < 2).
\end{equation}

The Tweedie family can model the compound structure of ``a point mass at zero + continuous positive values'' under a unified distributional assumption, which better matches the real GMV generation process. Meanwhile, the redemption sub-task (Redem head) uses cross-entropy loss. The overall objective consists of three terms: seller-level GMV supervision, redemption classification, and platform-level aggregation. Here, $\widehat{GMV}_{u,s_i}$ denotes the seller-level GMV prediction constructed by the seller-view or RDD branch, $\widehat{GMV}^{\text{pla}}_{u,s_i}$ denotes the platform-view prediction produced by the Platform head, and $\hat{r}_{u,s_i}$ denotes the redemption probability predicted by the Redem head:

\begin{equation}
\mathcal{L}_{\text{seller}} =
\sum_{u}\sum_{s_i\in\mathcal{S}_u}
\mathcal{L}_{\text{Tw}}\left(GMV_{u,s_i}, \widehat{GMV}_{u,s_i}\right),
\end{equation}

\begin{equation}
\mathcal{L}_{\text{pla}} =
\sum_{u}
\mathcal{L}_{\text{Tw}}\left(GMV^{\text{pla}}_{u}, \sum_{s_i\in\mathcal{S}_u} \widehat{GMV}^{\text{pla}}_{u,s_i}\right),
\end{equation}

\begin{equation}
\mathcal{L}_{\text{redem}} =
\sum_{u}\sum_{s_i\in\mathcal{S}_u}
\mathrm{CE}\left(r_{u,s_i}, \hat{r}_{u,s_i}\right).
\end{equation}

The final training objective is:
\begin{equation}
\mathcal{L}_{\text{total}} =
\mathcal{L}_{\text{seller}}
+ \lambda_{\text{pla}}\mathcal{L}_{\text{pla}}
+ \lambda_{\text{redem}}\mathcal{L}_{\text{redem}}.
\end{equation}

In our experiments, $\lambda_{\text{pla}}$ and $\lambda_{\text{redem}}$ are set to 1 unless otherwise specified. The overall training procedure is summarized in Appendix~\ref{sec:appendix_training}.

\section{Experiment}

\subsection{Datasets}

\subsubsection{Industrial Dataset}
Our industrial data are collected from the online logs of a large-scale e-commerce system. This dataset constitutes a highly challenging task: when aggregating the Individual Treatment Effects (ITE) for all associated sellers of a single user using baseline models, the estimated total gain overestimates the actual online observed effect by approximately 35\%. Furthermore, among the behaviors where coupons are issued and conversions occur, approximately 25\% ultimately fail to complete redemption, indicating significant ``pseudo-conversion'' noise that further interferes with the estimation of true incremental effects.

\subsubsection{Synthetic Dataset}

We construct a synthetic dataset comprising user features $\mathbf{x}_u$, seller features $\mathbf{x}_s$, treatment variable $t$, and potential outcomes $y_0$ and $y_1$. To simulate seller cannibalization, we introduce a decay mechanism where uplift diminishes based on the number of same-category sellers the user has sequentially encountered. Due to the complexity of multi-incentive interactions and compliance heterogeneity, we focus solely on seller-level cannibalization in synthetic data, while RDD's effectiveness on incentive cannibalization is validated through industrial data ablations. Complete generation details are provided in Appendix~\ref{sec:appendix_synthetic}.

\subsection{Evaluation Metrics}

\subsubsection{AUUC}
AUUC is computed based on the Uplift curve, which depicts: when we rank the population from high to low according to the model's predicted uplift scores and select different proportions of the population for intervention, the cumulative total uplift of the treatment group relative to the control group within that subset\cite{gutierrez2017causal}. The value at any point $\text{Uplift}(\phi)$ on the curve is computed as:
\begin{equation}
\text{Uplift}(\phi) = \left( \frac{R^T(\phi)}{N^T(\phi)} - \frac{R^C(\phi)}{N^C(\phi)} \right) \times (N^T(\phi) + N^C(\phi)),
\end{equation}
where: $\phi$ represents the population proportion selected after ranking by the model's uplift scores from high to low (e.g., $\phi=0.1$ represents the top 10\% with the highest scores); $N^T(\phi)$ and $R^T(\phi)$ are the total number of users and the total GMV generated in the treatment group within that population, respectively; $N^C(\phi)$ and $R^C(\phi)$ correspond to the control group.

AUUC measures the area under the Uplift curve:
\begin{equation}
\text{AUUC} = \int_{0}^{1} \text{Uplift}(\phi) \, d\phi, 
\end{equation}
where a larger value indicates stronger model capability in identifying high-uplift users.

\subsubsection{QINI}
The QINI curve depicts the cumulative net incremental gain at each population proportion $\phi$. The value at any point $Q(\phi)$ is computed as:
\begin{equation}
Q(\phi) = \sum_{i \in T(\phi)} y_i - \left( \sum_{j \in C(\phi)} y_j \right) \cdot \frac{|T(\phi)|}{|C(\phi)|}.
\end{equation}
The core idea is to calibrate for differences in treatment and control group sizes, thereby fairly evaluating net gain.

The QINI coefficient is defined as the area between the model's QINI curve and the random selection baseline:
\begin{equation}
\text{QINI} = \int_0^1 (Q_{\text{model}}(\phi) - Q_{\text{random}}(\phi)) \, d\phi.
\end{equation}
A higher QINI value indicates better model performance in identifying individuals with high treatment effects.

\subsubsection{Weighted AUUC/QINI}
In the industrial dataset with multiple treatment types (e.g., coupons with different values), we adopt weighted AUUC and QINI (wAUUC and wQINI). Continuous treatment values are discretized into bins, with AUUC/QINI computed per bin and aggregated via sample-size weighting.

We report AUUC/QINI at two granularities. Seller-level metrics evaluates uplift ranking for each user--seller candidate pair, reflecting the model's ability to identify high-increment seller-level allocation opportunities. User-level metrics first aggregates the predicted uplift over candidate sellers associated with the same user and then evaluates the ranking at the user/platform level, reflecting whether the model can identify users with higher platform-wide incremental potential. For the industrial dataset, both seller-level and user-level metrics are computed in their weighted forms due to multiple treatment values.

\begin{table*}[htbp]
\centering
\caption{Main experimental results on both synthetic and industrial datasets.}
\label{tab:main_results}
\begin{tabular}{lccccccccc}
\toprule
\textbf{Model} & \multicolumn{4}{c}{\textbf{Industrial Dataset}} & \multicolumn{4}{c}{\textbf{Synthetic Dataset}} \\
\cmidrule(lr){2-5} \cmidrule(lr){6-9}
& Seller AUUC & Seller QINI & User AUUC & User QINI 
& Seller AUUC & Seller QINI & User AUUC & User QINI \\
\midrule
DragonNet & 0.650 & 0.248 & 0.705 & 0.238 & 1.0475 & 0.5432 & 0.9438 & 0.4430 \\
CFRNet    & 0.694 & 0.261 & 0.748 & 0.276 & 1.0084 & 0.4966 & 0.9960 & 0.4839 \\
TARNet    & 0.729 & 0.284 & 0.771 & 0.294 & 1.0074 & 0.5030 & 1.1052 & 0.5972 \\
TLearner  & 0.730 & 0.289 & 0.782 & 0.301 & 0.8308 & 0.3305 & 0.9548 & 0.4512 \\
EUEN      & \underline{0.744} & \underline{0.297} & \underline{0.794} & \underline{0.319} & 1.0940 & 0.5837 & \underline{1.1245} & \underline{0.6024} \\
X-net     & 0.713 & 0.273 & 0.786 & 0.307 & 1.0612 & 0.5589 & 0.9481 & 0.4553 \\
M3TN      & 0.709 & 0.281 & 0.775 & 0.313 & \underline{1.1382} & \underline{0.6273} & 0.9714 & 0.4591 \\
UMLC      & 0.681 & 0.255 & 0.742 & 0.271 & 1.0291 & 0.5263 & 0.9165 & 0.4238 \\
HUM       & 0.701 & 0.266 & 0.759 & 0.289 & 0.9714 & 0.4728 & 0.9612 & 0.4613 \\
\textbf{Ours}      & \textbf{0.769} & \textbf{0.314} & \textbf{0.849} & \textbf{0.348} & \textbf{1.1596} & \textbf{0.6380} & \textbf{1.1474} & \textbf{0.6237} \\
\bottomrule
\end{tabular}
\end{table*}

\subsection{Experimental Setup}
\label{sec:exp_setup}

\textbf{Hardware.} All experiments on the industrial dataset are conducted on the internal production system. Experiments on the synthetic dataset are also performed on the internal computing platform with abundant computational resources.

\textbf{Optimization and Training.} We employ Optuna\cite{akiba2019optuna} for automated hyperparameter tuning with 40 trials per experiment. Each trial trains for a maximum of 30 epochs with early stopping if validation metrics do not improve for 5 consecutive epochs. The batch size is set to 4096 with L2 regularization coefficient $\lambda = 1 \times 10^{-4}$. The search space includes learning rate $\eta \in \{5 \times 10^{-4}, 1 \times 10^{-4}, 5 \times 10^{-5}\}$ and hidden dimension $d_h \in \{128, 256, 512\}$.

\subsection{Main Experimental Results}

We select DragonNet\cite{shi2019adapting}, CFRNet\cite{shalit2017estimating}, TARNet\cite{shalit2017estimating}, TLearner\cite{kunzel2019metalearners}, EUEN\cite{ke2021addressing}, M3TN\cite{m3tn}, UMLC\cite{sun2025robust}, HUM\cite{zhai2026heterogeneous} as baseline methods, with evaluation metrics using the commonly adopted AUUC and QINI in uplift modeling. As shown in Table~\ref{tab:main_results}, on the industrial dataset, DragonNet, CFRNet, and TARNet, which adopt shared network architectures with multi-head structures, perform poorly. In contrast, TLearner and EUEN, which decouple the control network independently, demonstrate clear advantages. This is primarily because, unlike simple binary treatment problems, our business scenario contains very rich treatment-related features, and the control group prediction should shield this information to avoid bias; this decoupling is more easily achieved in the EUEN and TLearner structures. Since EUEN performs best among all baseline models, we select it as our baseline.

In the industrial dataset, ``Ours'' denotes the addition of Platform-level Global Alignment and Redemption-based Decomposition Denoising to the EUEN baseline, abbreviated as PGA and RDD, respectively. In the synthetic dataset, since only seller-level cannibalization is simulated, ``Ours'' denotes the addition of PGA to EUEN. Across both datasets and all metrics, our method achieves the best performance. The synthetic results verify the effectiveness of PGA under controlled seller-level cannibalization, while the industrial results further demonstrate the benefit of jointly mitigating seller-level and incentive-level cannibalization in real-world deployment.

\subsection{Ablation Studies}

To verify the effectiveness of each module, we conduct ablation experiments. The baseline uses EUEN to estimate GMV-related uplift. As shown in Table~\ref{tab:ablation}, adding PGA improves performance. Since RDD uses redemption labels as supervision, we introduce a control variant, denoted as ``+Redem'', which adds redemption prediction as an auxiliary task but does not decompose the treated outcome into redemption and non-redemption paths. This comparison isolates the benefit of the proposed path-level decomposition from the mere use of redemption supervision. Under the same redemption-label supervision, RDD still improves over +Redem, showing that the decomposition structure contributes beyond auxiliary redemption prediction. When PGA and RDD are combined, they mutually reinforce each other, yielding the best performance.

It is noteworthy that PGA brings a particularly large improvement on User AUUC, suggesting that platform-level alignment helps rank users by their platform-wide incremental potential. In contrast, RDD yields consistent gains over the +Redem variant, especially on QINI-related metrics, indicating that explicitly separating redemption and non-redemption paths helps reduce over-attribution in cumulative uplift estimation. AUUC mainly reflects ranking quality, while QINI emphasizes cumulative net gain. These results show that PGA and RDD improve uplift estimation from complementary perspectives: PGA aligns seller-level predictions with platform-level growth, while RDD reduces attribution noise from mixed conversion paths.

\begin{table}[htbp]
\centering
\caption{Ablation study results on the industrial dataset. We compare the base EUEN model with different combinations of Platform-level Global Alignment (PGA) and Redemption-based Decomposition Denoising (RDD).}
\label{tab:ablation}
\begin{tabular}{lccccc}
\toprule
\textbf{Metric} & \textbf{Baseline} & \textbf{+PGA} & \textbf{+Redem} & \textbf{+RDD} & \textbf{Ours} \\
\midrule
Seller AUUC & 0.744 & 0.751 & 0.749 & 0.757 & \textbf{0.769} \\
Seller QINI & 0.297 & 0.302 & 0.299 & 0.308 & \textbf{0.314} \\
User AUUC   & 0.794 & 0.826 & 0.803 & 0.818 & \textbf{0.849} \\
User QINI   & 0.319 & 0.337 & 0.326 & 0.332 & \textbf{0.348} \\
\bottomrule
\end{tabular}
\end{table}

\begin{figure}[htbp]
    \centering
    \includegraphics[width=0.45\textwidth]{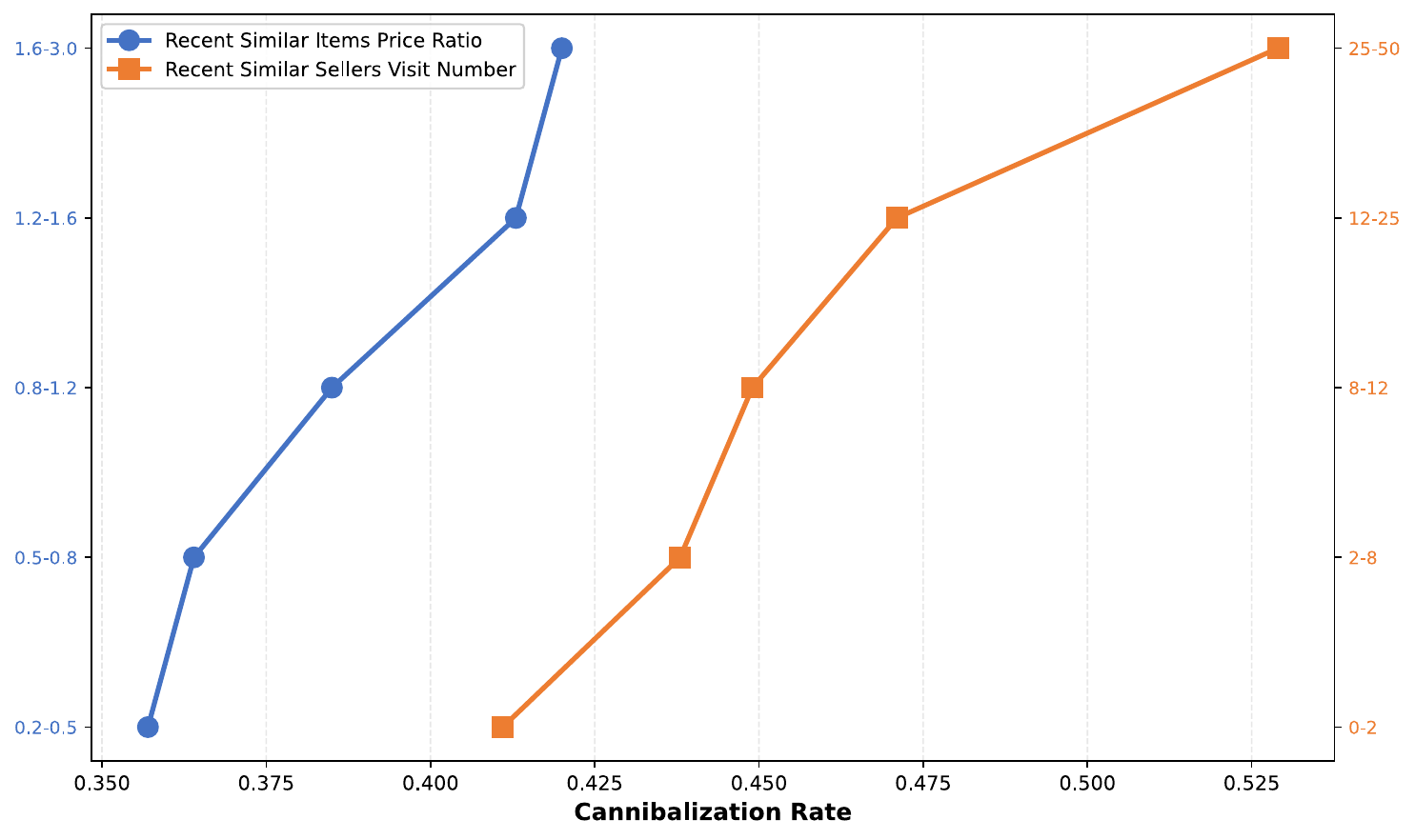}
    \caption{Correlation between estimated cannibalization rate and user behavioral features. X-axis: model-predicted cannibalization rate $g$. Blue curve: ratio of recently browsed same-category item prices to current item price. Orange curve: number of same-category sellers visited in the past 10 days. Both features show monotonic positive correlation with cannibalization rate.}
    \label{fig:cannibalization_analysis}
\end{figure}

\subsection{Seller Cannibalization Analysis}

To verify whether the model effectively learns cannibalization effects, we define the individual estimated cannibalization rate as:
\begin{equation}
g = \frac{p_{\text{seller}} - p_{\text{platform}}}{p_{\text{seller}}},
\end{equation}
where $p_{\text{seller}}$ is the predicted seller-level incremental GMV and $p_{\text{platform}}$ is the predicted platform-level incremental GMV.

Since individual-level ground truth cannibalization rates are unavailable, we validate the model's effectiveness by analyzing two typical cannibalization scenarios: (1) users who frequently visit same-category sellers recently, indicating strong organic purchase intent where coupons may fail to generate true incrementality; (2) users whose recently browsed same-category items have significantly higher average prices than the current item, where coupons may lead users to purchase cheaper items, resulting in negative incrementality. Based on these hypotheses, we conduct validation using users' click sequences from the past 10 days.

As shown in Fig.~\ref{fig:cannibalization_analysis}, both behavioral features demonstrate strong positive correlations with estimated cannibalization rates, validating our modeling assumptions. The blue curve shows that users who recently browsed higher-priced same-category items exhibit higher cannibalization rates, confirming that coupons for lower-priced items are more likely to divert purchases from higher-value alternatives. The orange curve reveals that users who visited more same-category sellers also show elevated cannibalization rates, indicating that cross-shop browsing behavior signals stronger organic intent and substitution effects. These consistent patterns verify that the PGA module effectively captures seller-level cannibalization signals during training.

\subsection{Incentive Cannibalization Analysis}

To validate the effectiveness of the Redemption-based Decomposition Denoising (RDD) module in mitigating incentive cannibalization, we analyze the predicted uplift for users who converted without redeeming their assigned coupons. These ``conversion without redemption'' cases represent a key source of noise, as they may be driven by organic demand or alternative incentives rather than the assigned treatment, leading to overestimated uplift.

\begin{figure}[htbp]
    \centering
    \includegraphics[width=0.45\textwidth]{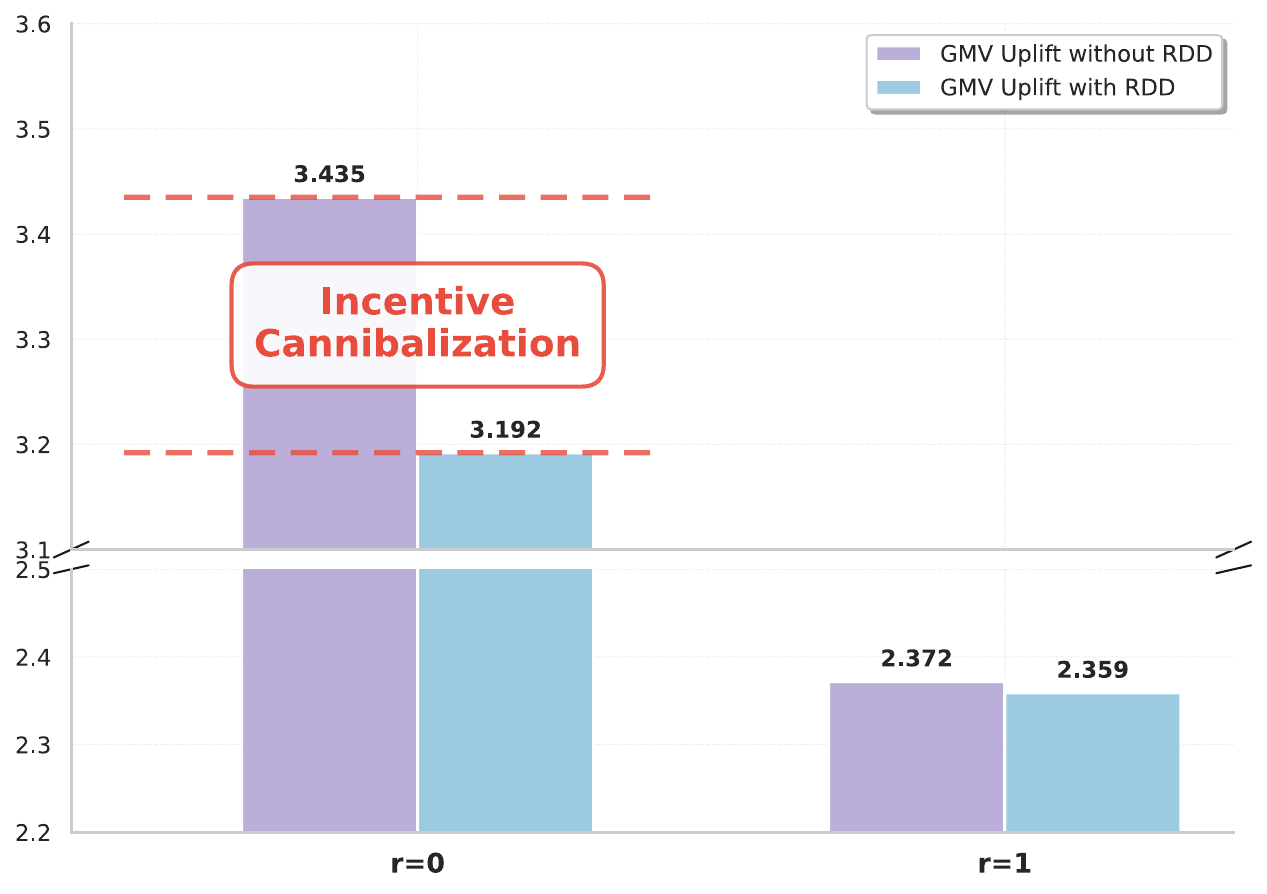}
    \caption{Comparison of predicted GMV uplift between baseline (without RDD) and proposed method (with RDD), segmented by redemption behavior. Left (r=0): users who converted without redeeming the assigned coupon. Right (r=1): users who converted with redemption. Numbers indicate average predicted uplift in each segment. The baseline overestimates uplift in the r=0 group by 7.6\% (3.435 vs 3.192), while both models align in the r=1 group.}
    \label{fig:rdd_effectiveness_compressed}
\end{figure}

Fig.~\ref{fig:rdd_effectiveness_compressed} compares the predicted uplift between the baseline model (without RDD) and our proposed method (with RDD) across two user groups segmented by redemption behavior (r). For users who redeemed their coupons ($r=1$), both models produce similar predictions, since redemption provides a direct observable signal that the assigned coupon participated in the conversion path. However, for users who converted without redemption (r=0), the baseline model predicts substantially higher uplift compared to our RDD-equipped model. This discrepancy reveals the presence of incentive cannibalization: the r=0 group likely includes users with strong organic intent or those influenced by higher-value concurrent incentives, leading the baseline to overattribute incrementality to the assigned (but unused) coupon. By explicitly decomposing the two conversion paths, RDD assigns lower uplift to such non-redemption conversions, reducing overestimation by approximately 7\% in the r=0 segment. This denoising pattern suggests that RDD mitigates over-attribution in non-redemption conversions and improves the robustness of uplift estimation.

\subsection{Online A/B Test Results}

\begin{table}[htbp]
\centering
\caption{Online A/B test results}
\label{tab:online_ab}
\begin{tabular}{ccc}
\toprule
Marketing Cost & Platform $\Delta$GMV & Platform ROI \\
\midrule
-2.45\% & +4.08\% & +6.69\% \\
\bottomrule
\end{tabular}
\end{table}

To validate the effectiveness of the proposed method in real-world business scenarios, we deployed an online A/B test from October 5, 2025 to October 12, 2025. ROI is defined as:
\begin{equation}
\text{ROI} = \frac{\Delta \text{GMV}}{\text{Cost}},
\end{equation}
where $\Delta$GMV represents the incremental GMV of the treatment group relative to the control group. 

As shown in Table~\ref{tab:online_ab}, compared to the production baseline model, our approach reduces marketing costs by 2.45\% and improves platform-wide $\Delta$GMV by 4.08\%, resulting in a 6.69\% increase in platform-wide ROI. These results confirm that our framework effectively denoises cannibalization effects in ITE estimation and reduces opportunity cost waste, demonstrating significant practical value in real-world business applications.

\section{Related Work}
\subsection{Foundations of Uplift Modeling}
Uplift modeling, or Individual Treatment Effect estimation, is rooted in the Potential Outcomes Framework\cite{rubin1974estimating, holland1986statistics} and Causal Diagram theory\cite{pearl2009causality}. Early research focused on Tree-based methods, which adapt splitting criteria to maximize the difference in treatment effects\cite{hansotia2002incremental, rzepakowski2010decision, athey2016recursive}. Subsequently, Meta-learners gained popularity for their flexibility in using any supervised learning algorithm as a base. This category includes S-Learner\cite{kunzel2019metalearners}, T-Learner\cite{kunzel2019metalearners}, X-Learner\cite{kunzel2019metalearners}, and R-Learner\cite{nie2021quasi}, which utilize different objective functions to minimize the mean squared error of the estimated uplift. While foundational, these methods often struggle with high-dimensional features and selection bias in observational data\cite{imbens2015causal, wager2018estimation}.

\subsection{Deep Representation Learning for ITE}
With the success of deep learning, neural network-based approaches have been proposed to learn balanced representations to mitigate selection bias\cite{johansson2016learning, louizos2017causal, yao2018representation}. A seminal work, CFRNet\cite{shalit2017estimating}, introduced the Integral Probability Metric (IPM) to regularize the distance between treatment and control group distributions\cite{schwab2018perfect, johansson2016learning}. To further enhance estimation accuracy, DragonNet\cite{shi2019adapting} incorporated propensity score estimation into representation learning, leveraging the sufficiency of the propensity score for unbiasedness\cite{rosenbaum1983central, assaad2021counterfactual}. Recent advances also explore information-theoretic bounds\cite{curth2021inductive} and generative adversarial networks (GANs)\cite{yoon2018ganite} to synthesize counterfactual outcomes. However, these models predominantly operate under the SUTVA assumption, which ignores the complex interdependencies in competitive marketplaces.

\subsection{Multi-treatment and Multi-task Learning in Marketing}
E-commerce platforms often involve multiple treatment options (e.g., various coupon types or discount levels), leading to the development of multi-treatment uplift models\cite{zhao2017uplift, acharki2023comparison}. Multi-task learning (MTL) architectures, such as MMoE\cite{ma2018modeling} and PLE\cite{tang2020progressive}, have been adapted to jointly model conversion and redemption\cite{zhao2019recommending,jin2022multi}. For instance, EUEN\cite{ke2021addressing} and DESCN\cite{zhong2022descn} modeling the entire user space to alleviate data sparsity. Despite these efforts, existing MTL-based uplift models typically treat individual seller increments as independent goals, failing to account for the Seller-level Cannibalization where gains in one shop are offset by losses in another.

\subsection{Causal Inference under Interference and Noise}
The violation of SUTVA, known as interference or spillover effects, has been studied in social networks and marketplace experiments\cite{tchetgen2012causal,johari2022experimental}. In e-commerce, this manifests as the cannibalization of organic traffic or cross-seller competition. Furthermore, the gap between treatment assignment and actual redemption—what we term Incentive Cannibalization—introduces significant measurement noise\cite{wen2020entire}. While recent works have explored denoising techniques for recommender systems\cite{wang2021denoising, guo2025causal}, their application in uplift modeling remains sparse. Our work, CanniUplift, bridges this gap by proposing a holistic framework that explicitly models multi-source cannibalization through platform-level alignment and redemption-based denoising.

\section{Conclusion}

In this paper, we address the critical yet often overlooked challenge of multi-source cannibalization in e-commerce uplift modeling. By relaxing the SUTVA assumption, we identify two primary forms of interference: Seller Cannibalization, where incentives merely shift expenditure between shops, and Incentive Cannibalization, where rewards are redeemed by organically motivated users. To this end, we propose CanniUplift, which effectively filters out ``pseudo-incrementality'' by enforcing global consistency and explicitly modeling the redemption mechanism. Extensive experiments on both synthetic and large-scale industrial datasets demonstrate that CanniUplift significantly outperforms state-of-the-art uplift models. More importantly, online A/B tests in a real-world production environment confirm that our framework can reduce marketing costs while simultaneously driving higher platform-wide incremental GMV and ROI. These results underscore the importance of accounting for cross-unit interference in complex marketplace ecosystems.

\section{Limitation}

CanniUplift explicitly models seller-level and incentive-level cannibalization but does not address temporal cannibalization, where promotions cause users to shift future purchases forward and inflate short-term uplift at the cost of long-term revenue. Additionally, because CanniUplift relies on historical user behavior to estimate cannibalization effects, it suffers from latency in capturing real-time substitution. Future work will explore sequential and forward-looking uplift modeling, such as predicting future seller visits and coupon redemptions, to better capture delayed purchase shifts and dynamic substitution patterns.

\bibliographystyle{ACM-Reference-Format}
\bibliography{main}


\appendix

\section{Synthetic Dataset Generation}\label{sec:appendix_synthetic}

\subsection{Data Configuration}
The synthetic dataset includes user features $\mathbf{x}_u \in \mathbb{R}^{106}$, seller features $\mathbf{x}_s \in \mathbb{R}^{107}$, treatment variable $t \in \{0,1\}$, and potential outcomes $\{y_0, y_1\}$. We explicitly model seller-level cannibalization effects to simulate diminishing marginal returns under repeated exposure across similar seller segments.

\subsection{Feature Generation}

\paragraph{User Features.}
User features comprise $p=106$ dimensions: $p_b=23$ binary features sampled from $\text{Bernoulli}(0.5)$, $p_c=83$ continuous features from $\mathcal{N}(0,1)$, and one normalized categorical attribute derived from 10 discrete categories, represented as a continuous value in $[0,1)$.

\paragraph{Seller Features.}
Seller features comprise $q=107$ dimensions: $q_c=106$ continuous features from $\mathcal{N}(0,1)$, and one categorical feature with 300 original categories (normalized to $[0,1)$). This categorical feature is mapped to $K=10$ semantic segments via a Gaussian kernel-based soft clustering mechanism to model heterogeneous treatment effects. An auxiliary cluster label
\begin{equation}
x_{\text{cluster}} \sim \text{Multinomial}\left(6; \left[\frac{1}{6}, \dots, \frac{1}{6}\right]\right)
\end{equation}
is introduced for response generation only.

\subsection{User–Seller Interaction Structure}
We generate 500,000 unique sellers. For each of the 20,000 users, we randomly assign exactly $N=50$ sellers with replacement. Samples from the same user share identical $\mathbf{x}_u$ and treatment assignment $t$.

\subsection{Treatment Assignment}
Following the Randomized Controlled Trial (RCT) paradigm, treatment is assigned at the user level:
\begin{equation}
t \sim \text{Bernoulli}(0.5)
\end{equation}
All seller samples for a given user inherit the same $t$ value.

\subsection{Response Generation}

\paragraph{Control Outcome.}
The control outcome is defined as:
\begin{equation}
y_0 = a \cdot \|\mathbf{x}_u\|_2^2 + b \cdot \|\mathbf{x}_s\|_2^2 + c \cdot \phi(\mathbf{x}_u^\top \mathbf{x}_s) + d \cdot x_s^{\text{class}} + z_0 + \epsilon_0
\end{equation}
where $\phi(s)$ is a truncated quadratic function:
\begin{equation}
\phi(s) = 
\begin{cases}
s^2, & |s| \leq \delta \\
2\delta|s| - \delta^2, & \text{otherwise}
\end{cases}
\quad (\delta = 15),
\end{equation}
$x_s^{\text{class}}$ is the normalized seller categorical attribute, $\epsilon_0 \sim \mathcal{N}(0, 1)$, and $z_0$ is a cluster-specific offset sampled based on $x_{\text{cluster}}$ from: $\mathcal{N}(0,1)$, $\mathcal{N}(2,0.5)$, $\mathcal{N}(-1,2)$, $\mathcal{N}(3,1.5)$, $\mathcal{N}(-2,0.8)$, or $\mathcal{N}(1,2)$. Coefficients are $(a, b, c, d) = (1, 1, 0.03, 0.5)$.

\paragraph{Treatment Outcome.}
The treatment outcome is:
\begin{equation}
y_1 = y_0 + e \cdot \mathbf{1}^\top \mathbf{x}_u + f \cdot \mathbf{1}^\top \mathbf{x}_s + g \cdot (\mathbf{x}_u^\top \mathbf{x}_s) + z_1 + \epsilon_1 - \Delta
\end{equation}
where $(e, f, g) = (2, 2, 0.2)$, $\epsilon_1 \sim \mathcal{N}(0, 10)$, $z_1$ follows the same distribution as $z_0$, and $\Delta \geq 0$ is the gain decay term reflecting diminishing marginal returns from repeated exposure.

\subsection{Seller-Level Cannibalization Mechanism}

Existing synthetic data generation methods for uplift modeling, such as UMLC\cite{sun2025robust}, typically assume independent seller effects and ignore cross-seller interference. In contrast, we propose a novel cannibalization-aware simulation mechanism that explicitly models diminishing marginal returns when users are repeatedly exposed to sellers within the same semantic segment.

\paragraph{Soft Clustering.}
The 300 original seller categories are soft-clustered into $K=10$ semantic segments via Gaussian kernels. Given category $c \in [0, 299]$, the probability of mapping to segment $s \in [0, 9]$ is:
\begin{equation}
P(s|c) = \frac{\exp\left(-\frac{(c - \mu_s)^2}{2\sigma^2}\right)}{\sum_{k=0}^{9} \exp\left(-\frac{(c - \mu_k)^2}{2\sigma^2}\right)}
\end{equation}
where $\mu_s = 15 + 30s$ and $\sigma = 8$.

\paragraph{Uplift Decay Rule.}
For the $i$-th seller of user $u$ with segment $s_i$, let $n_s^{<i}$ denote the number of occurrences of segment $s$ among the first $i-1$ sellers. The uplift decay is:
\begin{equation}
\text{uplift}_i^{\text{new}} = \text{uplift}_i^{\text{raw}} - \min\left(d \cdot n_s^{<i}, 0.5 \cdot |\text{uplift}_i^{\text{raw}}|\right) \cdot \text{sgn}(\text{uplift}_i^{\text{raw}})
\end{equation}
where $d = 4 + 2(n_s^{<i} - 1)$ when $n_s^{<i} \geq 1$, otherwise $d=0$. This design ensures that repeated exposure to the same segment leads to cumulative decay, capped at 50\% of the original uplift magnitude.

\begin{table*}[hbtp]
\centering
\caption{Training Procedure of CanniUplift}
\label{alg:canniuplift_training}
\begin{tabular}{p{0.96\textwidth}}
\toprule
\textbf{Input:} Training data $\mathcal{D}$ with user features, candidate sellers,
seller-level GMV/redemption labels, and platform-level GMV. \\
\midrule
1. For each user $u$, encode multi-source behaviors:
$H_u, h_u \leftarrow \mathrm{Encoder}(x_u)$. \\

2. For each candidate seller $s \in \mathcal{S}_u$ of user $u$, compute candidate interaction:
$z_{u,s} \leftarrow \mathrm{TreatAttn}(s, H_u)$. \\

3. Predict control-side GMV:
$pGMV_c(s) \leftarrow \mathrm{CtrlHead}(z_{u,s})$. \\

4. Predict platform-view GMV:
$pGMV^{\mathrm{pla}}_s \leftarrow \mathrm{PlaHead}(z_{u,s})$. \\

5. Predict redemption probability:
$p_r(s) \leftarrow \mathrm{RedemHead}(z_{u,s})$. \\

6. Predict path-specific increments:
$p\Delta GMV_r(s), p\Delta GMV_{1-r}(s)
\leftarrow \mathrm{RDDHeads}(z_{u,s})$. \\

7. Combine the RDD prediction:
$pGMV_s \leftarrow pGMV_c(s)
+ p_r(s)p\Delta GMV_r(s)
+ (1-p_r(s))p\Delta GMV_{1-r}(s)$. \\

8. Aggregate platform prediction:
$pGMV^{\mathrm{pla}}_u \leftarrow
\sum_{s\in\mathcal{S}_u} pGMV^{\mathrm{pla}}_s$. \\

9. Compute $\mathcal{L}_{\mathrm{seller}}$,
$\mathcal{L}_{\mathrm{pla}}$, and $\mathcal{L}_{\mathrm{redem}}$. \\

10. Update parameters by minimizing $\mathcal{L}_{\mathrm{total}}$. \\
\bottomrule
\end{tabular}
\end{table*}

\section{Training Procedure of CanniUplift}
\label{sec:appendix_training}

Table~\ref{alg:canniuplift_training} summarizes the training procedure of CanniUplift, where candidate-specific representations are computed through Treat-Attention and the seller-level, platform-level, and redemption-related heads are jointly optimized using the objective defined in Section~\ref{sec:loss}.

\section{Inference with Platform-level Global Alignment}\label{sec:appendix_inference}

\subsection{From Global Training to Marginal Inference}

During training, the Platform head enforces global consistency through an aggregated loss over all candidate sellers:

\begin{equation}
\mathcal{L}_{\text{pla}} = \sum_u \mathcal{L}\left(\sum_{s_i \in \mathcal{S}_u} pGMV^{\text{pla}}_{s_i}, GMV^{\text{pla}}_u\right)
\end{equation}

However, at inference time, our objective is to evaluate the \textbf{marginal impact on platform-wide GMV} when changing the intervention status of a single seller $s_j$ from control to treatment, while keeping all other sellers' states unchanged.

\subsection{Practical Approximation for Marginal Scoring}

Consider a user $u$ with candidate seller set $\mathcal{S}_u$. The platform-level marginal increment induced by treating seller $s_j$ is defined as:

\begin{equation}
\begin{aligned}
\Delta_{\text{Platform}}(u, s_j) &= \mathbb{E}[Y^{\text{pla}}_u \mid t_j{=}1] - \mathbb{E}[Y^{\text{pla}}_u \mid t_j{=}0] \\
&= \sum_{s \in \mathcal{S}_u} \mathbb{E}[y_{u,s} \mid t_j{=}1] - \sum_{s \in \mathcal{S}_u} \mathbb{E}[y_{u,s} \mid t_j{=}0]
\end{aligned}
\end{equation}

For practical deployment, we need to estimate the marginal platform-level effect of assigning a coupon to a target seller $s_j$ for user $u$. A full counterfactual comparison would require recomputing the platform-wide GMV under two worlds, i.e., assigning and not assigning the treatment to $s_j$. For sellers $s_k \neq s_j$, their treatment states are fixed as background interventions during this comparison. However, their outcomes may still be affected by changing $t_j$, which is exactly the seller-level interference effect considered in this work. Exhaustively recomputing the counterfactual outcomes of all other sellers is computationally expensive at serving time. Therefore, we adopt a practical marginal scoring approximation: the Platform head is trained under the global aggregation constraint above, so its candidate-specific platform-view output is encouraged to absorb the expected net impact of treating $s_j$ on the user's total platform GMV.

Concretely, we compute two platform-view predictions for the target seller $s_j$:
\begin{equation}
\begin{aligned}
p_{\mathrm{treat}} &= pGMV^{\mathrm{pla}}_{s_j}(u,s_j,t_j{=}1), \\
p_{\mathrm{control}} &= pGMV^{\mathrm{pla}}_{s_j}(u,s_j,t_j{=}0).
\end{aligned}
\end{equation}

The platform-level marginal uplift is then approximated by their difference:
\begin{equation}
\Delta_{\mathrm{pla}}(u,s_j)
\approx p_{\mathrm{treat}} - p_{\mathrm{control}}
= \delta_{\mathrm{platform}}(u,s_j).
\end{equation}

This approximation should not be interpreted as assuming that other sellers' outcomes are unaffected. Instead, cross-seller substitution effects are implicitly captured through the parameters learned from the platform-level aggregation loss and the candidate-specific user--seller representation.

\subsection{Why Single-Point Predictions Capture Global Effects}

The key insight is that the global aggregation constraint during training \textbf{encourages the model to encode cross-seller substitution effects} into the model parameters. Specifically:

\begin{itemize}
\item \textbf{Global loss propagates cross-seller dependencies}: When the model overestimates $pGMV^{\text{pla}}_{s_i}$ for one seller during training, causing $\sum_{s \in \mathcal{S}_u} pGMV^{\text{pla}}_s$ to deviate from the true platform GMV, the gradient simultaneously corrects predictions for all related sellers.

\item \textbf{User representation captures substitution patterns}: The Treat-Attention mechanism introduced in Section~\ref{sec:3-1} enables the user behavior sequence $H_u$ to capture cross-shop browsing patterns (e.g., ``user alternates between similar-category sellers''). The model learns to infer global consumption elasticity from $\langle user, seller \rangle$ features.

\item \textbf{Inference as a counterfactual query}: The single-point score $\delta_{\text{platform}}(u, s_j)$ effectively answers: ``If we assign a coupon to user $u$ at seller $s_j$, how will their \textbf{total platform spending} change?'' Since the model has learned global dependencies, this score naturally reflects the net increment accounting for cannibalization.
\end{itemize}

\subsection{Practical Inference Procedure}

At inference time, for each user $u$ and candidate seller $s_j$:

\begin{enumerate}
\item Compute platform-view prediction with treatment:\\
$p_{\text{treat}} = \text{PlatformHead}(z_{u,s_j}, t_j{=}1)$

\item Compute platform-view prediction without treatment:\\
$p_{\text{control}} = \text{PlatformHead}(z_{u,s_j}, t_j{=}0)$

\item Calculate marginal platform uplift:\\
$\Delta_{\text{platform}}(u, s_j) = p_{\text{treat}} - p_{\text{control}}$

\item Rank sellers by $\Delta_{\text{platform}}$ and allocate top-K coupons
\end{enumerate}

\textbf{Crucially}, inference requires only single-point predictions for the target seller $s_j$, \textbf{without} aggregating over other sellers in $\mathcal S_u$. This design significantly reduces computational cost while preserving cannibalization-aware estimation.

\end{document}